\documentclass[runningheads]{llncs}
\usepackage[T1]{fontenc}
\usepackage{graphicx}
\usepackage[ruled, vlined, linesnumbered]{algorithm2e}
\usepackage{algorithmic}
\usepackage{latexsym}
\usepackage{amssymb}
\usepackage{amsfonts}
\usepackage{comment}
\usepackage{caption}
\usepackage{multirow}
\usepackage{amsmath}
\usepackage{cite}
\usepackage{amsmath,amssymb,amsfonts}
\usepackage{algorithmic}
\usepackage{graphicx}
\usepackage{textcomp}
\usepackage{xcolor}
\usepackage{multirow}
\usepackage{color}
\usepackage{epsfig}
\usepackage{url}
\usepackage{subfigure}

\begin{document}
\title{Predicting Parking Lot Availability by Graph-to-Sequence Model: A Case Study with SmartSantander}
%
%
\author{Yuya Sasaki\inst{1} \and
Junya Takayama\inst{1} \and
Juan Ramón Santana\inst{1} \and
Shohei Yamasaki\inst{1} \and
Tomoya Okuno\inst{1} \and
Makoto Onizuka\inst{1}
}
%
%
\institute{Osaka university \email{sasaki@ist.osaka-u.ac.jp}}
\maketitle              
\begin{abstract}
Nowadays, so as to improve services and urban areas livability, multiple smart city initiatives are being carried out throughout the world. SmartSantander is a smart city project in Santander, Spain, which has relied on wireless sensor network technologies to deploy heterogeneous sensors within the city to measure multiple parameters, including outdoor parking information. In this paper, we study the prediction of parking lot availability using historical data from more than 300 outdoor parking sensors with SmartSantander. We design a graph-to-sequence model to capture the periodical fluctuation and geographical proximity of parking lots. For developing and evaluating our model, we use a 3-year dataset of parking lot availability in the city of Santander. Our model achieves a high accuracy compared with existing sequence-to-sequence models, which is accurate enough to provide a parking information service in the city. 
We apply our model to a smartphone application to be widely used by citizens and tourists.

\end{abstract}

\section{Introduction}

Smart city projects use information and communication technologies to obtain real-time data to manage their services more effectively.
Santander, Spain, started its smart city project called SmartSantander\footnote{http://www.smartsantander.eu/}, which envisioned a large-scale deployment of more than 12,000 Internet of Things (IoT) sensors such as traffic, environmental, and parking sensors~\cite{sanchez2014smartsantander, sasaki2021survey, sasaki2021smart,sasaki2020sequenced}. The aim of such deployment is to improve the city livability and support tourism in Santander.

One of the services provided in Santander is guiding drivers to available parking lots.
Thanks to the outdoor parking information gathered from the SmartSantander deployment, drivers are informed about available parking lots through two means. First, ten parking panels, deployed at the entrance to streets in the city center, show the number of available parking lots in the city center as well as each of the streets where they are deployed. Second, several applications provide this information remotely. Hence, drivers can plan whether to take their own cars or use the public transportation services based on the real-time parking information.

\begin{figure}[ttt]
 	\centering
 	\includegraphics[width=0.9\linewidth]{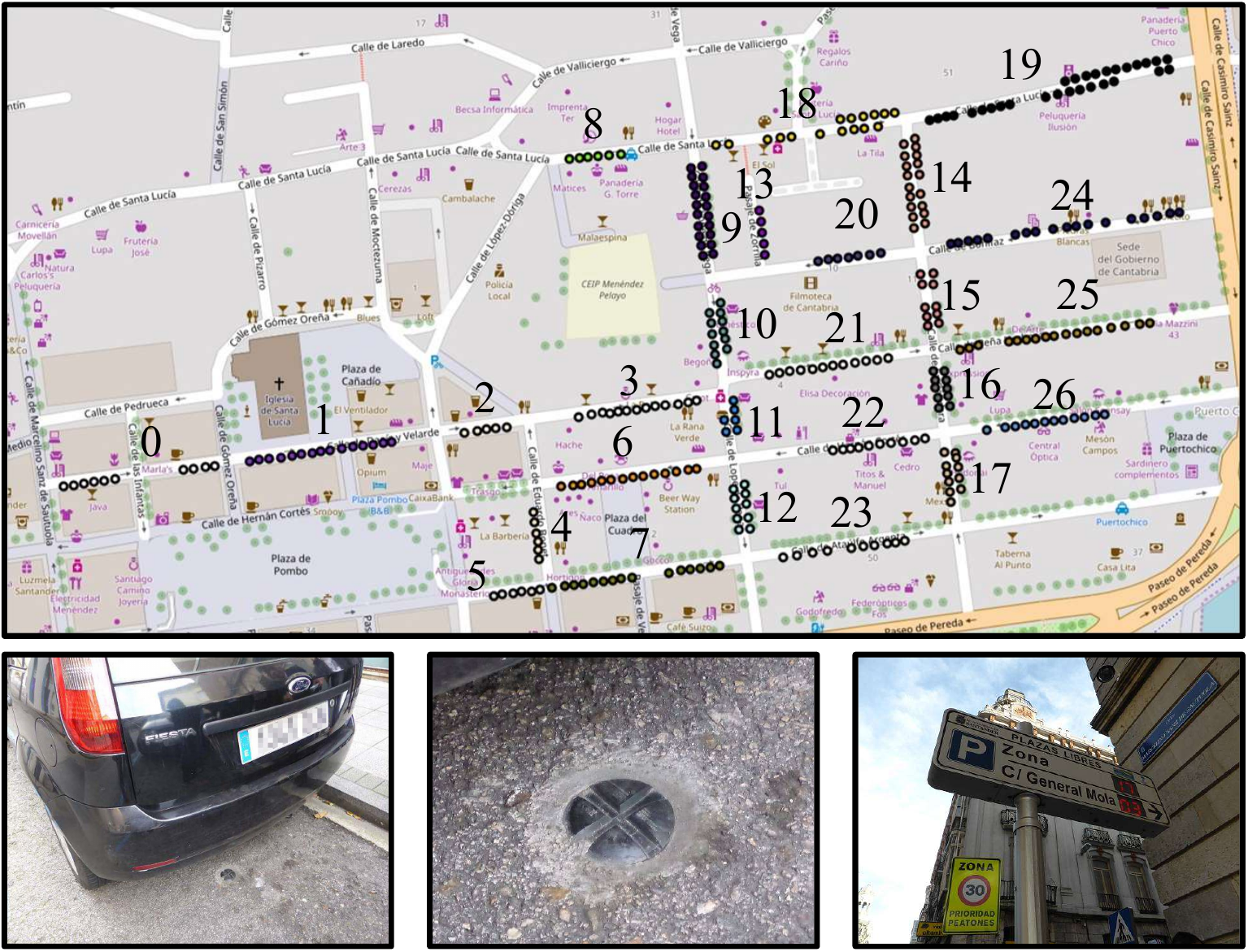}
 	\caption{Parking sensors in Santander. (top) locations of parking lots with cluster IDs, (bottom-left) a parking sensor with a car, (bottom-center) the parking sensor, and (bottom-right) a parking panel to guide drivers.}
 	\label{fig:intro}
 \end{figure}

Figure \ref{fig:intro} shows the deployment carried out in Santander, including the location of parking lots equipped with sensors, parking sensors, and the parking panels deployed in the city. Such services providing parking lot availability improve the efficiency and usefulness of the parking system.

{\bf Motivation: }
Despite the benefits provided by this deployment, SmartSantander is limited to offer real-time information to the users about available parking lots. A prediction service based on historical data can benefit the system deployed in Santander.
Such prediction service would allow drivers to know whether there are available parking lots or not beforehand, even if the service has been accessed long before the arrival to the parking lot.
Furthermore, SmartSantander can keep offering parking guiding service by predicting the number of available parking lots based on historical data, even if the service stops due to several issues, such as deployment maintenance or urban works.

There are two requirements to start a service that predicts parking availability in Santander: (1) predicting the number of available parking lots for both, the whole city center (i.e., all parking lots) and each of the streets; and (2) predicting their availability at multiple time steps without pre-defined time steps (e.g., in 15, 30, and 60 minutes, and more).
To the best of our knowledge, there is no former work that predicts the availability of parking lots per street at several time steps.

{\bf Contribution: } In this paper, we study the problem on the prediction of parking lot availability with SmartSantander.
To this end, we design a graph-to-sequence neural network model that can predict parking lot availability at multiple time steps by using historical data from parking lots.
Our graph-to-sequence model consist of two components: encoder and decoder. 
For the encoder, we develop a graph neural network (GNN) encoder to capture the characteristics of parking lot availability: (1) temporal, as parking lot availability trends change periodically and (2) spatial, as the availability of each parking lot is also affected by nearby parking lots.
Our decoder is based on a recurrent neural network (RNN) that uses time information (e.g. a day) as well as historical parking lot availability data to learn the impact of holidays and weekdays.

We demonstrate that our model is able to accurately predict the number of available parking lots in periods of 15, 30, 60 and 120 minutes. 
Our model outperforms existing sequence-to-sequence models.
We also develop a smartphone application with our prediction model to provide accurate prediction to citizens and tourists.


{\bf Reproducability:} We open our source code in Github and parking data under requests\footnote{\url{https://github.com/yuya-s/SatanderParking}}.

{\bf Organization:} The remainder of this paper is organized as follows. 
Section \ref{sec:dataset} describes the parking sensors in Satander.
Section \ref{sec:proposal} presents our approach, and then Section \ref{sec:experiment} shows the results obtained from the experiments.
Section \ref{sec:application} shows our smartphone application with the prediction model.
Section \ref{sec:related} introduces the related work. Finally, Section \ref{sec:conclusion} summarizes the paper and describes the future work.

\section{Parking sensors with SmartSantander}
\label{sec:dataset}


As aforementioned, we develop our prediction model for parking sensors deployed in Santander. There are 323 parking sensors manufactured by Nedap\footnote{\url{ https://nedap.com/en/}} and each sensor corresponds to a single parking lot. All these sensors are connected to a wireless network based on the standard IEEE802.15.4. 
The parking deployment follows a three-layered architecture composed of the following elements:
\begin{itemize}
\item Parking sensors: they are buried under the asphalt and monitor the status of the parking lots by measuring the change of the magnetic field on top of them. If the magnetic field measurement exceeds the calibrated threshold, they send a data frame with the new status to the closest relay node. 
\item Relay nodes: they are installed in the lampposts and building facades. They provide coverage to parking sensors and forward data frames to the data collector. 
\item Data collectors: they forward the information received from relay nodes to the central servers where data is stored, by using wide area network technologies (e.g. 3G, fiber ring). 
\end{itemize}



Monitored parking lots are located in a special area in the city center, in which the drivers have to pay park fee from 10:00 to 14:00 and 16:00 to 20:00 during weekdays and Saturday morning. Parking time is restricted to two hours per vehicle, except for those citizens who live nearby and have a special permission. Therefore, the status of parking lots changes frequently in these periods. Parking lots show a high occupancy for the full day, even over night.


 \section{Prediction model}
 \label{sec:proposal}
 
In this section, we first explain the requirements and problem definition that we solve in this paper. Then, we explain the preprocessing for the parking data. Finally, we explain our graph-to-sequence neural network model and its training method.

\subsection{Requirements and Problem Definition}


As we described in Section 1, we have two requirements to start the prediction services in Santander.
First, we need to predict the number of available parking lots for both, the whole city center (i.e., entire parking area) and per street.
While, we do not need to predict the status of each parking lot because the parking services in Santander provides street-based parking availability.
Second, we need to predict the number of available parking lots at multiple time steps instead of a single time step. These time steps should not be defined in advance, because people may move from further areas to the city center by their cars and the distances are unsure.

We here formally define the problem we solve in this paper.
We have the set $P$ of historical parking lot data for each parking lot.
$p_i \in P$ is a vector whose size is the number of parking lots and it consists of 0/1 values, where 1 and 0 represent whether each parking lot is available or not at time step $i$, respectively.
We define that $\langle p_{i}, \ldots, p_{j} \rangle$ is a sequence of vectors from time step $i$ to $j$.
\smallskip
\\\noindent
{\bf Problem Definition: }
{\it Given the set of historical parking lots data $P$, we build a model to accurately predict the parking lot availability. In this model, given a sequence of vectors $\langle p_{t-M}, \ldots, p_{t-1} \rangle$ as input, it outputs the number of available parking lots per street from time step $t$ to $t'$ ($t'$ is not given in advance).}

\subsection{Preprocessing}

As preprocessing of parking lots data, we cluster the parking lots and construct a graph  based on the closeness among parking lots .

\smallskip
\noindent
{\bf Clustering.}
It is effective because nearby parking lots are likely to have a similar behavior, as we do not have specific parking lots but specific areas where we can park our cars.
We cluster the parking lots per street, considering that this is how SmartSantander provides the available parking lots to citizens. After this process, we obtain 27 different clusters. Figure \ref{fig:intro}(top) shows these clusters, where different colors represent each one of them. The number of parking lots differs for each of the clusters. For instance, clusters 2, 9, and 19 in Figure \ref{fig:intro}(top) include 5, 20, and 29 parking lots, respectively.

We normalize the values in the vector dividing them by the number of parking lots in the cluster in order to reduce unnecessary effects of the clusters that include large numbers of parking lots.
Finally, we have the set of vectors $S$ that contains a vector $s_i$ whose size is 27 and values are from 0 to 1 at time step $i$.
We input $\langle s_{t-M}, \ldots, s_{t-1} \rangle$ to models, and the models output $\langle s_{t}, \ldots, s_{t'} \rangle$.

\smallskip
\noindent
{\bf Graph construction.} We build a graph whose vertices are clusters of parking lots and two vertices have an edge if the parking lots represented by the two vertices are closer than a given spatial threshold.
We assume that parking lots represented by the two vertices are affected each other if both vertices have edges. 
So as to locate each cluster, we use the centroid of the locations of parking lots that belong to that cluster. Two vertices have edges if the Euclidean distance between them is 95 meters, which is decided based on the length of streets.

\subsection{Graph-to-Sequence model}

Our graph-to-sequence model consists of a {\it GNN encoder} and an {\it RNN decoder}. 
The GNN encoder maps the input to a vector of a fixed dimensionality by graph neural networks, and the RNN decoder receives and decodes it to generate the sequence of predicted parking lots availability. 

\smallskip
\noindent
{\bf GNN encoder model.}
Our GNN encoder model captures the temporal perspective by 1D-convolution and the spatial perspective by graph convolution, which captures local trends of nearby parking lots in stead of global trends.

\begin{figure*}[ttt]
 	\centering
 	\includegraphics[width=1.0\linewidth]{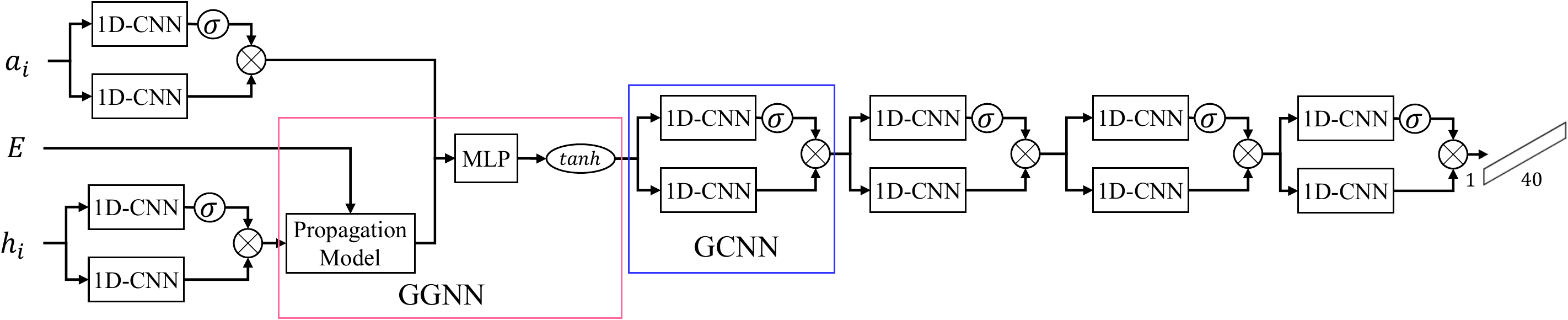}
 	\caption{GNN encoder model. Our GNN encoder consists of one GGNN and six GCNN layers.}
 	\label{fig:gnn}
 \end{figure*}

\begin{figure*}[ttt]
 	\centering
 	\includegraphics[width=1.0\linewidth]{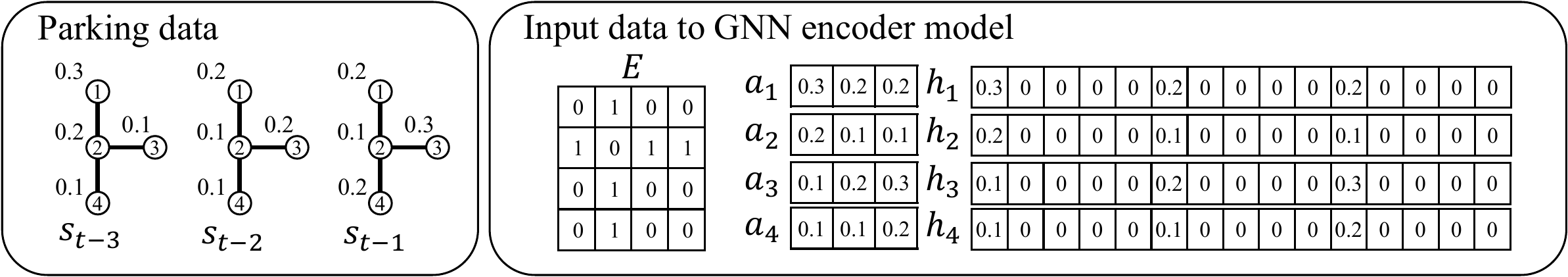}
 	\caption{Input data to the GNN encoder model}
 	\label{fig:gnninput}
 \end{figure*}

Figures \ref{fig:gnn} and \ref{fig:gnninput} show our GNN encoder model and the input data for this model, respectively. 
Our GNN encoder model combines gated graph neural networks (GGNN for short) \cite{li2015gated} with gated convolutional neural network (GCNN for short) \cite{dauphin2017language}.
The GGNN takes graph convolution to capture spatial perspective and the GCNN takes 1D-convolution to capture temporal perspective.
Propagation model in the GGNN conducts graph convolution that learns about the effect from neighbor clusters (i.e., vertices connected by edges).
Consecutive GCNNs after the GGNN emphasizes the characteristics by gradually reducing the size of the vector.
Our GNN encoder has three input parameters, parking  $a_i$, hidden state $h_i$, and set of edges $E$ of the graph.
$a_i$ is a vector whose values are the number of available parking lots of cluster $i$ at $M$ steps, and $h_i$ is a vector that is extended from $a_i$ and the extended area has zero values. 

In the following, we describe our GNN encoder model in detail.
$[;]$ and $\otimes$ denote concatenation and element-wise multiplication, respectively .



Our GNN encoder model first applies GCNNs to $a_i$ and $h_i$, respectively.
The GCNNs have two 1D-convolutional layers; the top 1D-CNN computes the importance via a sigmoid function and the bottom one computes values themselves. This is the same structures of the original GCNNs.
$a_i'$ and $h_i'$ are outputs of the GCNNs, which are calculated as follows:
\begin{eqnarray}
\label{eq:gate1_a}
a_i' = (a_i \ast \Gamma_{a,f} + b_{a,f}) \otimes \sigma (a_i \ast \Gamma_{a,g} + b_{a,g})\\
\label{eq:gate1_h}
b_i' = (b_i \ast \Gamma_{h,f} + b_{h,f}) \otimes \sigma (h_i \ast \Gamma_{h,g} + b_{h,g})
\end{eqnarray}
where $\sigma$ denotes sigmoid function.
$\Gamma_{a,f}, \Gamma_{a,g}$ and $\Gamma_{h,f}, \Gamma_{h,g}$ are 1-dimensional convolution operators for annotations and hidden states, respectively.
$b_{a,f}, b_{a,g}$ and $b_{h,f}, b_{h,g}$ are correspondence biases.

The GGNN aggregates hidden states of neighbors at each time step (Eq. (\ref{eq:message})), and then updates the hidden states like GRU (Eqs. (\ref{eq:GRU_R})--(\ref{eq:GRU_H}))~\cite{GRU}. Then, an output of message passing is applied to MLP layer (Eq.~(\ref{eq:ggnnmlp})).
The propagation model aggregates and updates repeatedly by the following equations:

\begin{eqnarray}
h{(0)} &=& h_i'\\
\label{eq:message}
X_{(j)} &=& E [h_{(j-1)}W_{1}+b_{1};\dots;h_{(j-1)}W_{M}+b_{M}]\\
\label{eq:GRU_R}
r_{(j)} &=& \sigma(M_{(j)}W_{r} + h_{(j-1)}U_{r})\\
z_{(j)} &=& \sigma(M_{(j)}W_{z} + h_{(j-1)}U_{z})\\
\widehat{h_{(j)}} &=& \tanh(M_{(j)}W_h+(h_{(j-1)} \otimes r_{(j)}U_h))\\
\label{eq:GRU_H}
h_{(j)} &=& (1-z_{(j)}) \otimes h_{(j-1)} + z_{(j)} \otimes \widehat{h_{(j)}}
\end{eqnarray}


\noindent
where $j$ denotes the time of iterations and $X, r, z$ denote the message, reset gate, and update gate, respectively. $W_{1},\dots,W_{M},W_{r}, U_{r}, W_z, U_z, W_h, U_h $ and $b_{1}$, $\dots$, $b_{M}$ are learnable parameters.
The parameters $W_{1}$, $\dots$, $W_{M}$, $b_{1}$, $\dots$, $b_{M}$ are for message passing.

We apply an output model to the $I$ times-propagated hidden state $h_{(I)}$.
An output $O_g$ of the GGNN is calculated as follows:
\begin{equation}\label{eq:ggnnmlp}
O_g = \tanh([h_{(I)};a_i']W_{O} + b_{O}) 
\end{equation}
\noindent
where $W_{O}, b_{O}$ are learnable parameters.
We then repeatedly apply four GCNNs to $O_g$ that equations are the same as Eq. (\ref{eq:gate1_a}).

We explain parameters of our GNN encoder in detail.
In the 1D-convolutional layers for $a_i$, the number of filters, filter size, and stride length are 1, 2, and 1, respectively.
In the 1D-convolutional layers for $h_i$, the number of filters, filter size, and stride length are 2, 10, and 5, respectively.
Four consecutive 1D-CNNs after the GGNN have 4, 115, 60, and 65 as the filter sizes, 2, 5, 5, and $null$ as the stride length, and 5, 10, 20, and 40 as the number of filters, respectively.
We use $null$ as the stride length, as the filter is of the same size as the matrix size.
Finally, our GNN encoder outputs a vector whose size is 40, which is the input for our decoder model.

Our GNN can capture spatial perspective for every parking lot because we input the data for each single parking lot one by one along with the graph that represents the closeness of parking lots.
Therefore, our GNN model captures local spatial perspective effectively.



\smallskip
\noindent
{\bf RNN Decoder.}
Our RNN decoder employs a multilayered Long Short-Term Memory (LSTM)~\cite{hochreiter1997long} to output the sequence of predicted parking lots availability without pre-defined time steps.
We use time information $\boldmath{d}_t$ as input for the decoder in order to capture the periodical fluctuation. 
$\boldmath{d}_t$ includes four values that represent the time, day, month, and weekday.
Due to the use of time information in the decoder, our model effectively handles the periodical fluctuation, such as weekdays and weekends.

In our RNN decoder, each function is computed by the same equations of basic LSTM~\cite{hochreiter1997long}, with the exception that we use a ReLU function for embedding the inputs, and a sigmoid function for the output.
In more concretely, our decoder model consists of the follow equations:
\begin{eqnarray}
\hat{\boldmath {s}}_t &=& ReLU (W_s  \boldmath {s}_t + \boldmath {b}_s)\label{eq:LSTMinput1}\\
\hat{\boldmath {d}}_t &=& ReLU (W_d  \boldmath {d}_t + \boldmath {b}_d)\label{eq:LSTMinput2}\\
\boldmath {f}_t &=& \sigma (W_f[\boldmath {h}_{t};\hat{\boldmath{s}}_t;\hat{\boldmath {d}}_t]+\boldmath {b}_f)\nonumber\\
\boldmath {l}_t &=& \sigma (W_i[\boldmath {h}_{t};\hat{\boldmath{s}}_t;\hat{\boldmath {d}}_t]+\boldmath {b}_l)\nonumber\\
\tilde{\boldmath {C}}_t &=& \tanh (W_C[h_{t},\hat{\boldmath {s}}_t,\hat{\boldmath {d}}_t]+\boldmath {b}_C)\nonumber\\
\boldmath {C}_{t+1} &=& \boldmath {f}_t \otimes \boldmath {C}_{t} + \boldmath {l}_t  \tilde{\boldmath {C}}_t\nonumber\\
\boldmath {o}_t &=& \sigma (W_o[\boldmath{h}_{t};\hat{\boldmath {s}}_t;\hat{\boldmath {d}}_t]+\boldmath {b}_o)\nonumber\\
\boldmath {h}_{t+1} &=& \boldmath {o}_t \otimes \tanh \boldmath {C}_{t+1}\nonumber\\
\boldmath {s}_{t+1} &=& \sigma (W_{s'}  \boldmath {C}_{t+1} + \boldmath {b}_{s'}) \label{eq:LSTMend}
\end{eqnarray}
These equations are the same for basic LSTM cells except for Eqs. (\ref{eq:LSTMinput1})--(\ref{eq:LSTMend}).
Eqs. (\ref{eq:LSTMinput1}) and (\ref{eq:LSTMinput2}) are for embedding, and Eq. (\ref{eq:LSTMend}) is for output. 
Both $h_t$ and $C_{t}$ are initially set to the output of encoder.


\subsection{Training}
We use the mean absolute error (MAE) as the measure to quantify errors for our training data. Hence, we minimize the MAE measure in the objective function over training data.
Our loss function is defined as follows:
\begin{equation}
Loss = \frac{1}{N' \cdot |s|}\sum_{i=1}^{N'} \sum_{j=1}^{|s|} abs(s'_{i,j} - s_{i,j})
\end{equation}
where, $|s|$ denotes the size of vector $s$ (i.e., 27), and $s'_{i,j}$ and $s_{i,j}$ denote the predicted and measured values of cluster $j$ at time step $i$, respectively.
$N'$ is a predefined parameter for model training, and we set $N'$ to a random value.
To reduce overfitting, we apply dropout with a probability of 0.3 to the encoding.

\section{Experiment}
 \label{sec:experiment}
 
In this section we evaluate our graph-to-sequence model to predict parking lot availability from Santander dataset.
 
\subsection{Setting}
We provide an overview of our experimental set up, including the dataset, competitors, and parameters. 

\begin{figure}[ttt]
 	\centering
 	\includegraphics[width=0.7\linewidth]{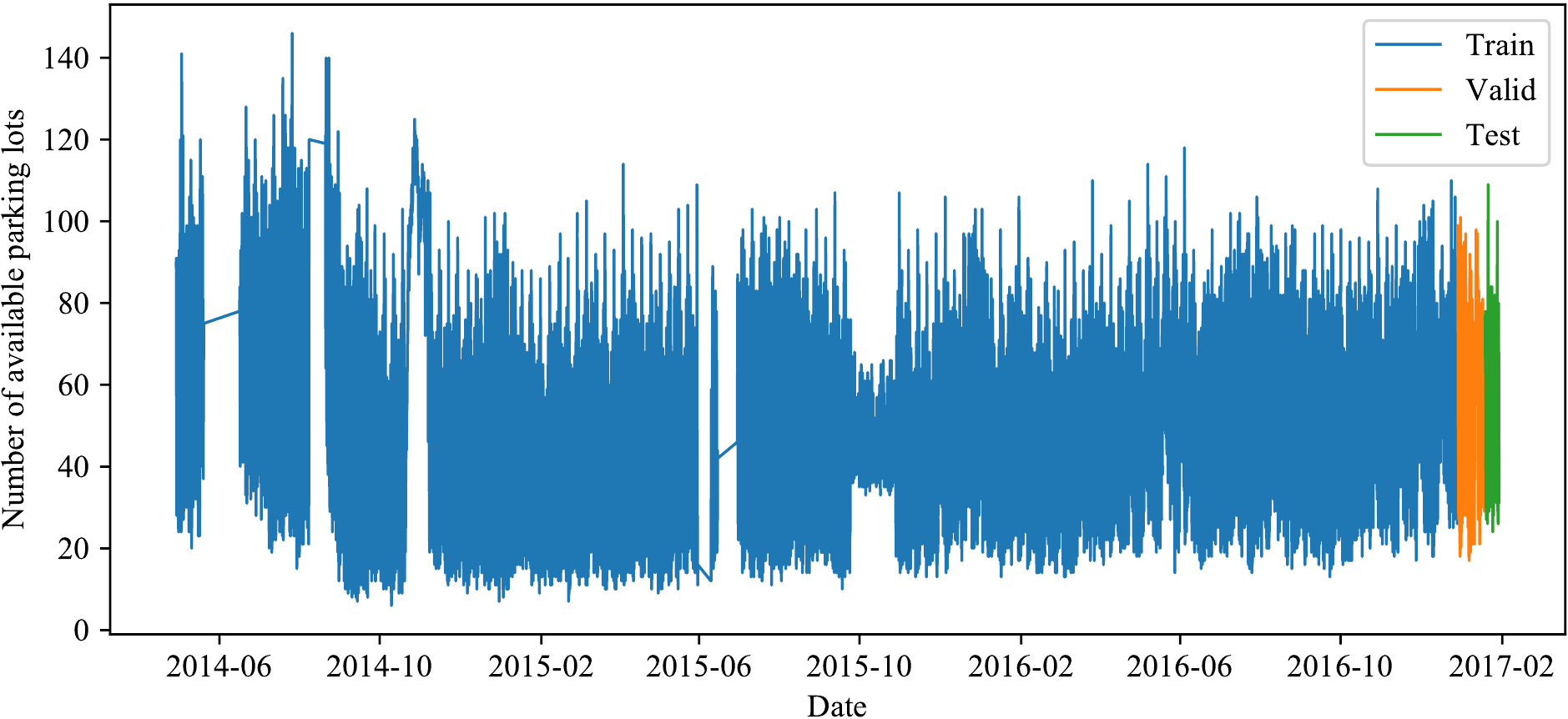}
 	\caption{Number of available parking lots}
 	\label{fig:prediciton}
 \end{figure}

\smallskip
\noindent
{\bf Dataset.}
We use a dataset with three years of parking data from Santander.
The number of parking lots covered in this dataset are 323, with data from 29th April 2014 to 29th January 2017.
We sample the data each 15 minutes.
Figures~\ref{fig:intro}(top) and \ref{fig:prediciton} show the location of parking lots and the number of available parking lots in the whole city, respectively. 
In Figure~\ref{fig:prediciton}, there are some periods in which there are few changes in the number of available parking lots are caused by maintenance.


\smallskip
\noindent
{\bf Competitors.}
We evaluate our graph-to-sequence model compared with two sequence-to-sequence models and two traditional auto-regressive (AR) models.

In sequence-to-sequence models, we use two types encoders RNN and CNN with the same decoder of our model.
The RNN encoder consists of LSTM as same as the decoder.
The CNN encoder convolutes a matrix containing $\langle s_{t-M}, \cdots, s_{t-1} \rangle$ to a fixed size vector.
We describe the detailed architectures of RNN and CNN encoders in a supplementary file.

In AR models, we use ARIMA and SARIMA \cite{box2015time}, two of the most widely used methods for time series forecasting.
ARIMA model assumes that the future value of a time series is a linear combination of historical values.
SARIMA model takes into account seasonality as well.


\smallskip
\noindent
{\bf Parameter.}
We divide the dataset into a training set from April 29th, 2014, to December 29th, 2016; a validation set from December 29th, 2016, to January 19th, 2017; and a test set from January 19th, 2017, to January 29th, 2017.
The input size of historical data $M$ is 48.
That is, we use parking data for 12 hours (i.e, 48 $\cdot$ 15 min) as historical data.
We predict the parking data in 15 min to 2 hours. During our preliminary experiments, we tested different input sizes for the model: 4, 24, 48, and 96 as $M$, from which the chosen input size achieved the high accuracy in average.
We also set the number $I$ of propagation in the GGNN to five based on our preliminary experiment.

When we train the model, we set the size of batch and epoch as 512 and 50, respectively.
Then, in the test phase we choose the same parameters that provide the highest accuracy in the validation phase averagely.

\begin{table}[t]
    \centering
    \caption{MAE of available parking lots in the entire parking area. The bold font indicates the best accuracy.}
    \label{tab:overview}
    \begin{tabular}{|c|cccc|}\hline
         & 15 min & 30 min & 60 min & 120 min\\ \hline\hline
        ARIMA &  15.31	&14.95	&15.415	&15.26\\
        SARIMA & 11.79	&12.04	&10.97	&11.49\\
        RNN &   10.95 & 11.18 & 11.16 & 12.13  \\
        CNN &   4.166  & 4.173  & 5.102  & 7.466 \\ 
        GNN &   {\bf 2.511}  & {\bf 3.301}  & {\bf 4.608}  & {\bf 7.242} \\ \hline
    \end{tabular}

\end{table}

\begin{table*}[t!]
\centering
\caption{MAE of available parking lots per street. The numbers within parentheses at Cluster ID denote the numbers of parking lots in clusters. The bold font indicates the best accuracy.}

\label{table:result}
\scalebox{0.8}{
\begin{tabular}{|l|llll|llll|llll|}\hline
\multirow{2}{*}{Cluster ID}            & \multicolumn{4}{c|}{RNN}       & \multicolumn{4}{c|}{CNN}& \multicolumn{4}{c|}{GNN}\\
  & 15 min & 30 min & 60 min & 120 min & 15 min & 30 min & 60 min & 120 min& 15 min & 30 min & 60 min & 120 min \\\hline\hline
0 (10)      & 0.749  & 0.734  & 0.769  & 0.769   & 0.468  & 0.497  & 0.540   & {\bf 0.606} & {\bf 0.293}&{\bf 0.392}&{\bf 0.502}&0.640
  \\
1 (15)      & 0.803  & 0.816  & 0.822  & 0.843   & 0.552  & 0.556  & 0.579  & 0.647 &   {\bf 0.301}&{\bf 0.367}&{\bf 0.472}&{\bf 0.576}  \\
2 (5)       & 0.404  & 0.392  & 0.416  & 0.429    & 0.088  & 0.095  & 0.117  & 0.155  &  {\bf 0.069}&{\bf 0.081}&{\bf 0.110}&{\bf 0.139}  \\
3 (13)      & 0.911  & 0.899  & 0.911  & 0.904   & 0.669  & 0.717  & 0.773  & 0.844  &  {\bf 0.243}&{\bf 0.319}&{\bf 0.444}&{\bf 0.606}  \\
4 (6)       & 0.265  & 0.291  & 0.302  & 0.301   & 0.126  & 0.148  & 0.175  & 0.231 &  {\bf 0.104}&{\bf 0.123}&{\bf 0.171}&{\bf 0.223} \\
5 (6)       & 1.015  & 0.978  & 0.968  & 0.975   & 0.513  & 0.530   & 0.587  & 0.632  &  {\bf 0.247}&{\bf 0.337}&{\bf 0.448}&{\bf 0.591}  \\
6 (14)      & 0.806  & 0.786  & 0.791  & 0.803   & 0.579  & 0.645  & 0.675  & 0.806 &  {\bf 0.284}&{\bf 0.379}&{\bf 0.495}&{\bf 0.673}  \\
7 (15)      & 0.913  & 0.936  & 0.966  & 1.042   & 0.510   & 0.561  & 0.599  & 0.738 &   {\bf 0.241}&{\bf 0.342}&{\bf 0.465}&{\bf 0.629}  \\
8 (6)       & 0.631  & 0.703  & 0.765  & 0.847   & 0.415  & 0.539  & 0.640   & 0.769 &   {\bf 0.243}&{\bf 0.368}&{\bf 0.524}&{\bf 0.673}  \\
9 (20)      & 1.550  & 1.585  & 1.611  & 1.666   & 0.589  & 0.653  & 0.715  & 0.817 &  {\bf 0.244}&{\bf 0.358}&{\bf 0.521}&{\bf 0.696}  \\
10 (12)     & 0.560  & 0.543  & 0.549  & 0.540   & 0.279  & 0.288  & 0.342  & 0.449 &   {\bf 0.131}&{\bf 0.184}&{\bf 0.244}&{\bf 0.327}  \\
11 (6)      & 0.422  & 0.405  & 0.401  & 0.417   & 0.240   & 0.261  & 0.281  & 0.321  &  {\bf 0.088}&{\bf 0.121}&{\bf 0.168}&{\bf 0.251}  \\
12 (10)     & 0.982  & 0.974  & 0.953  & 0.966   & 0.722  & 0.733  & 0.756  & 0.808 &  {\bf 0.301}&{\bf 0.400}&{\bf 0.539}&{\bf 0.699}  \\
13 (5)      & 0.804  & 0.846  & 0.874  & 0.948   & 0.178  & 0.204  & {\bf 0.256}  & 0.371  &  {\bf 0.127}&{\bf 0.179}& 0.265&{\bf 0.370}  \\
14 (18)     & 1.976  & 2.214  & 2.331  & 2.372   & 0.669  & 0.708  & 0.894  & 1.115  &  {\bf 0.248}&{\bf 0.361}&{\bf 0.554}&{\bf 0.806}  \\
    15 (10)     & 0.744  & 0.838  & 1.011  & 1.175   & 0.245  & 0.278  & 0.363  & 0.457 &  {\bf 0.131}&{\bf 0.170}&{\bf 0.252}&{\bf 0.339}  \\
16 (10)     & 0.806  & 0.886  & 0.912  & 1.001   & 0.314  & 0.331  & 0.399  & 0.490   &  {\bf 0.126}&{\bf 0.187}&{\bf 0.275}&{\bf 0.402}  \\
17 (11)     & 0.819  & 0.821  & 0.839  & 0.860   & 0.502  & 0.528  & 0.594  & 0.711  &  {\bf 0.229}&{\bf 0.309}&{\bf 0.399}&{\bf 0.586}  \\
18 (16)     & 1.203  & 1.197  & 1.193  & 1.167   & 0.701  & 0.725  & 0.758  & 0.821 &  {\bf 0.297}&{\bf 0.391}&{\bf 0.505}&{\bf 0.658}  \\
19 (29)     & 1.276  & 1.317  & 1.397  & 1.494   & 0.940   & 0.940   & 1.053  & {\bf 1.221}  &  {\bf 0.759}&{\bf 0.890}&{\bf 1.038}&1.238  \\
20 (7)      & 0.581  & 0.623  & 0.787  & 0.892   & 0.157  & 0.182  & 0.278  & 0.389  &  {\bf 0.066}&{\bf 0.108}&{\bf 0.212}&{\bf 0.351}  \\
21 (12)     & 0.796  & 0.824  & 0.834  & 0.850   & 0.524  & 0.526  & 0.593  & 0.680  &  {\bf 0.185}&{\bf 0.259}&{\bf 0.367}&{\bf 0.508}  \\
22 (10)     & 0.882  & 0.929  & 0.980  & 1.108   & 0.835  & 0.818  & {\bf 0.929}  & {\bf 1.088}  &  {\bf 0.580}&{\bf 0.737}& 0.944&1.211 \\
23 (10)     & 1.510  & 1.444  & 1.226  & 1.154   & 0.395  & 0.448  & 0.547  & 0.650   &  {\bf 0.191}&{\bf 0.295}&{\bf 0.429}&{\bf 0.572}  \\
24 (18)     & 1.352  & 1.165  & 1.004  & 1.036   & 0.572  & 0.653  & 0.736  & 0.907  &  {\bf 0.403}&{\bf 0.546}&{\bf 0.694}&{\bf 0.879}  \\
25 (17)     & 3.172  & 3.200  & 3.145  & 3.120   & 0.859  & 0.889  & 1.107  & 1.504  &  {\bf 0.346}&{\bf 0.509}&{\bf 0.816}&{\bf 1.258}  \\
26 (12)     & 0.925  & 0.956  & 0.985  & 1.057   & 0.567  & 0.628  & {\bf 0.702}  & {\bf 0.833}  &  {\bf 0.417}&{\bf 0.549}&0.707&0.929  \\\hline 
\end{tabular}
}
\end{table*}

\subsection{Experimental results}

\smallskip
\noindent
{\bf Comparing our models with baselines.}
We first compare prediction performance for our model and the four baselines. We evaluate MAE in 15, 30, 60, and 120 minutes time steps.
Table \ref{tab:overview} shows the MAE of the whole city for each time step and model.
The GNN and CNN encoders achieve the best and second best accuracy for all time steps, respectively.
The accuracy of RNN encoder is almost the same accuracy of the SARIMA model.
The GNN and CNN encoders capture the spatial perspective effectively.
While, RNN encoder and SARIMA are able to capture the seasonality effect, but they cannot capture spatial perspectives well.
From these results, we can confirm that spatial perspective are effective to parking prediction.
In addition, the parking lot prediction becomes more accurate when using nearby parking lots than the entire parking area because the GNN encoder achieves higher accuracy than the CNN encoder.


In our graph-to-sequence and sequence-to-sequence models, MAE increases as time steps increase, while in AR models the MAE value remains constant.
This is due to the fact that graph- and sequence-to-sequence models accumulate errors in the previous steps, thus the error becomes larger as time steps increase.


We evaluate prediction performance of GNN, CNN, and RNN encoders for the number of available parking lots per cluster.
Table \ref{table:result} shows the MAEs for each cluster. In the table, the numbers in brackets denote the number of parking lots in each of the clusters.
Since the MAEs of ARIMA and SARIMA are larger than those of our models (see Table~\ref{tab:overview}), we do not show these results.

The predicted errors are quite small even for the 120 minutes time step.
Since absolute errors are mostly less than one, each model often accurately predicts the number of available parking lots.
The GNN and CNN encoders outperform the RNN encoder model for all clusters at each time step as they are able to capture the spatial perspective more precisely than the RNN models. 
Comparing the GNN encoder with the CNN encoder, the GNN encoder achieves higher accuracy than the CNN encoder at most cases.

From these results, we can see that our graph-to-sequence model accurately predicts the number of available parking lots in the entire parking area and outperform baselines.





\begin{figure*}[ttt]
  \centering
  \subfigure[15 min]{\includegraphics[width=0.23\linewidth]{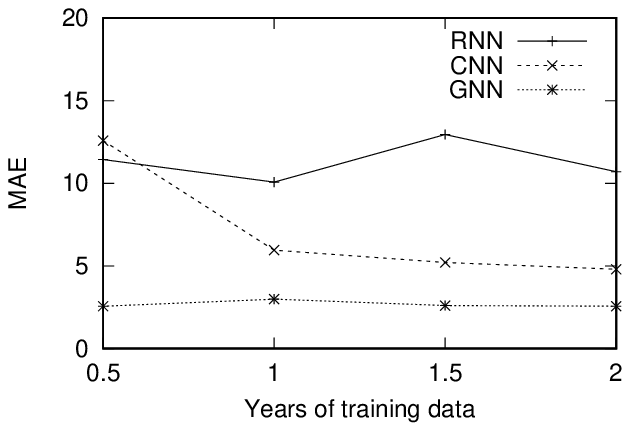}}
  \subfigure[30 min]{\includegraphics[width=0.23\linewidth]{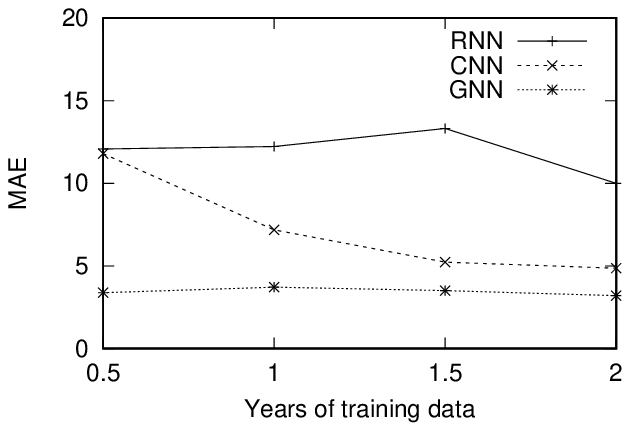}}
    \subfigure[60 min]{\includegraphics[width=0.23\linewidth]{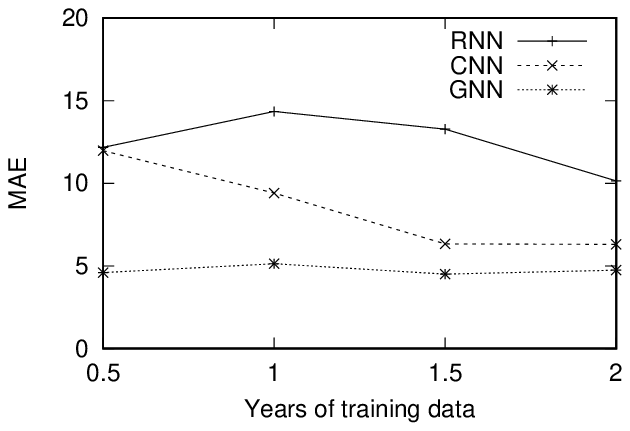}}
    \subfigure[120 min]{\includegraphics[width=0.23\linewidth]{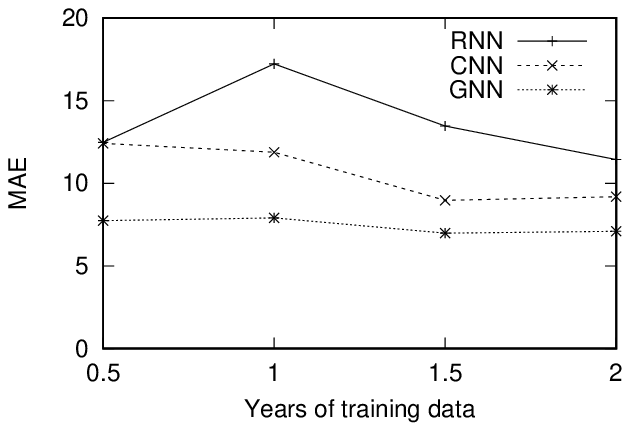}}
  \caption{Impact of training data sizes}
  \label{fig:trainingdata}
  \end{figure*}

\smallskip
\noindent
{\bf Impact of training data size.}
We evaluate the prediction performance varying the size of training data.
Figure~\ref{fig:trainingdata} shows the MAE using different training data sizes for all the time steps and models.
We use the same test and validation datasets, and the same date for the end of the training dataset. For instance, to obtain one year of training data, we train the models by data from December 29th, 2015 to December 29th, 2016.

From these results, the performance slightly changes for our GNN and the RNN encoders, even with smaller training datasets.
On the other hand, the prediction performance of the CNN encoder is lower when the training data size is only half of the years. 
This indicates that the CNN encoder needs a larger training dataset to be fine tuned.
Our GNN encoder always achieves the best performance among the models even with a small training dataset.
We can confirm that our GNN encoder captures temporal and spatial perspective more effectively than the other encoders.

 \begin{figure*}[!t]
 	\centering
 	\subfigure[RNN encoder]{
 	\includegraphics[width=0.95\linewidth]{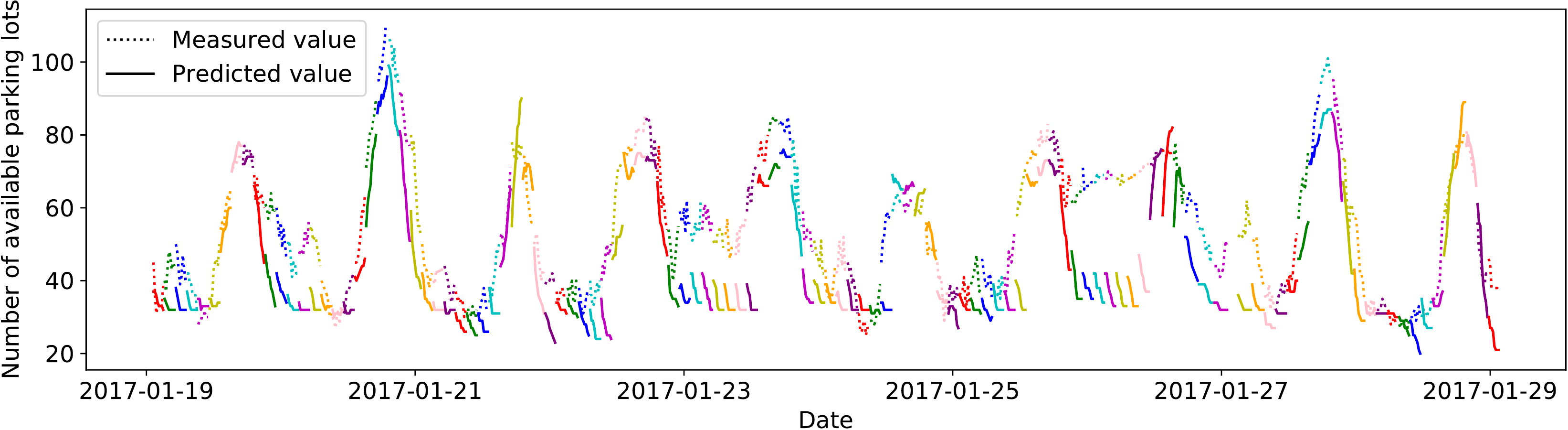}
 	}
 	 \subfigure[CNN encoder]{
 	\includegraphics[width=0.95\linewidth]{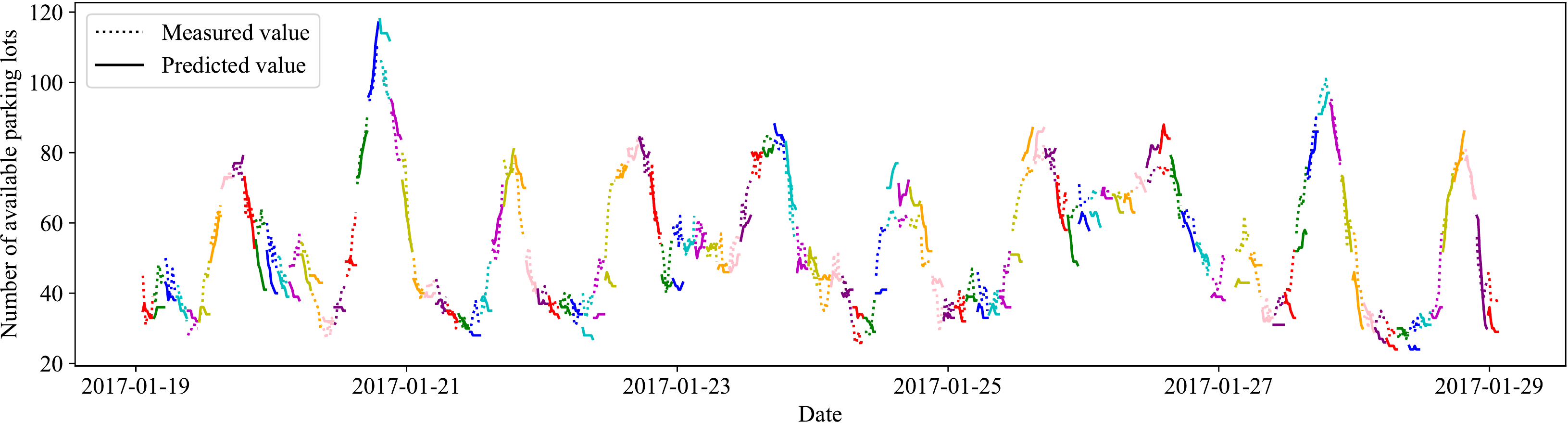}
 	}
 	\subfigure[GNN encoder]{
 	\includegraphics[width=0.95\linewidth]{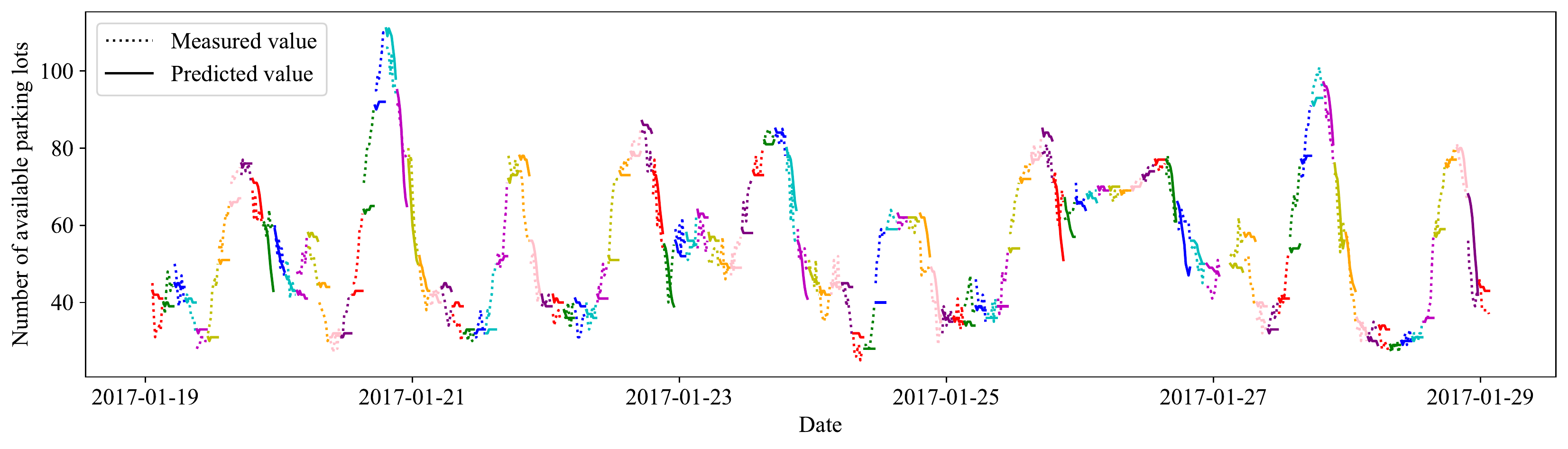}
 	}
 	\caption{Prediction performance at each time step. Each solid line represent two hours prediction. The solid and dashed lines with the same colors represent the same time periods.}
 	\label{fig:erroranalysis_timestep}
 \end{figure*}
 


 


\smallskip
\noindent
{\bf Error analysis at each time step.}
We finally analyse the transitions of the difference between measured and predicted values in GNN, CNN, and RNN encoders.
Figure \ref{fig:erroranalysis_timestep} shows the difference between measured and predicated values on RNN, CNN, and GNN encoders. 
Dashed lines represent measured values, while solid lines represent predicted values, and same colors represent same time period.
So, if dashed and solid lines with the same color are vertically close, prediction is accurate. We note that each solid line with each color indicates predicted values for eight steps (i.e., for 2 hours) and two neighbor solid lines with different colors are not overlapped.

From the results, we can see that the predicated values generally have a similar trend as measured values.
Figure \ref{fig:erroranalysis_timestep}(a) shows that the RNN encoder predictions  often underestimate available parking lots from measured values and have a large difference between predicted and measured values. In our use case, underestimated predictions are inadequate for citizens as they would be guided to streets with occupied parking lots.
Figure \ref{fig:erroranalysis_timestep}(b) shows that predicted values in the CNN encoder are closer to measured values than the RNN encoder.
However, the transition sometimes follows an opposite direction, for example, green and blue lines around 2017-01-26. 
In the case of the CNN encoder, we can notice from the figures that it cannot capture temporal perspective at some points.
Finally, Figure \ref{fig:erroranalysis_timestep}(c) shows that predicted values in the GNN encoder are also close to measured values.
The GNN enconder does not present opposite transitions. However, it consecutively outputs the same predicted values in several points. Hence, we can observe that our GNN encoder is conservative.

\section{Smartphone application}
\label{sec:application}
\begin{figure}[ttt]
 	\centering
 	\includegraphics[width=0.2\linewidth]{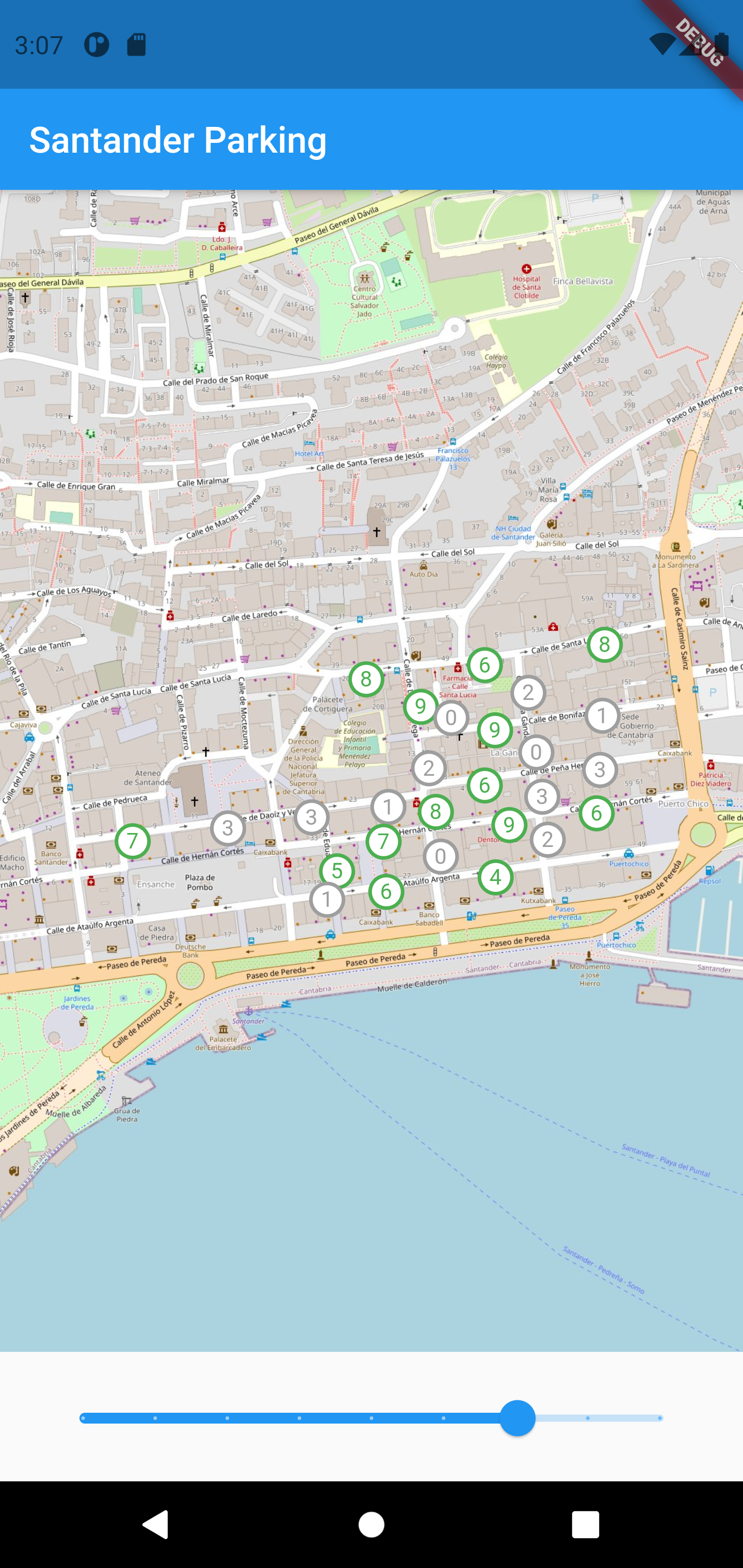}
	\caption{Screenshot of our smartphone application for parking prediction with SmartSantander}
 	\label{fig:iphone}
 \end{figure}

We apply our graph-to-sequence model to a smartphone application in order to provide accurate parking prediction to citizen and tourists.
Figure~\ref{fig:iphone} shows a screenshot of our smartphone application.
The application shows how many parking lots are currently available and will be available in 15--120 minutes.

We developed our application by Flutter for deploying it on both iOS and Andoroid platforms.
We run prediction models in a server, and then smartphones access to the server for obtaining and displaying the prediction results.
The server accesses the real-time parking information and the historical ones within 24 hours though APIs on data collector of SmartSantander.
This smartphone application is helpful to intuitively see the parking lots availability in the current and the future time.

\section{Related work}
\label{sec:related}

We review the similar existing works.
Since there are a large number of existing works on parking prediction, please refer a survey paper \cite{lin2017survey} that summarizes existing works related to smart parking systems.

{\bf Parking prediction.}
Due to the developments of smart city projects, nowadays many cities monitor the status of parking lots, such as Berlin \cite{tiedemann2015concept}, Barcelona \cite{caicedo2012prediction}, and Santander.
There are many works that aim to predict parking lot availability.
Tiedemann et al. \cite{tiedemann2015concept} developed a system that predicts occupancy for parking spaces in Berlin, Germany, which collects data from roadside parking sensors. This system predicts occupancy by using neural gas machine learning methods combined with data threads. 
Caicedo et al. \cite{caicedo2012prediction} developed a real-time availability forecast (RAF) algorithm for predicting real-time parking lot availability, located in Barcelona, Spain. 
Chen et al. \cite{chen2014parking} tackles the parking problem by aggregating parking lots in the same way as we do in our problem, in San Fransisco.
They evaluated multiple models, such as ARIMA, linear regression, support vector regression, and feed forward neural networks, and the neural network algorithm have achieved the best performance.

Several neural network-based prediction models have been proposed nowadays~\cite{yang2019deep,jomaa2019hybrid,shao2018parking,camero2018evolutionary,tekouabou2020improving}\footnote{These works are not coupled with smart city projects.}.
For example, Shao et al.~\cite{shao2018parking}, Jomaa et al.~\cite{jomaa2019hybrid}, Xiao et al.~ \cite{xiao2021hybrid} use LSTM, CNN, and GCN with GLU, respectively.
Existing works predict the availability of parking lots at either a single step (i.e., current or near future) or pre-defined multiple steps. As our aim is to predict availability of parking lots at non pre-defined multiple time steps simultaneously, existing algorithms are not applicable for our problem.
The sequence-to-sequence models that we used as our baselines can be considered as the extension of existing works \cite{shao2018parking,jomaa2019hybrid}, and we validated that our method achieves higher accuracy than them.


Some existing methods use not only historical parking data, but also other data sources for predicting parking lot availability.
For example, data sources include historical data generated by mobile phones (e.g., \cite{nandugudi2014pocketparker}), data extracted from vehicles equipped with GPS receivers (e.g., \cite{pflugler2016predicting}), and information from web maps (e.g., 
\cite{arora2019hard,yang2021truck,zhao2021mepark}).
In this regard, our model could be improved if we include these extra data. However, such data is difficult to be accessed, and we must deal with privacy related aspects, as well as to consider that not all the users use such recent devices and services.
So, they are not suitable in the situations of SmartSantander.

{\bf Deep learning on spatial and temporal data prediction.}
There are similar works with parking availability prediction such as predicting traffic density, pollution data, and the general availability of resources \cite{rottkamp2018time,jiang2021dl}.
These works are categorized into grid-based and graph-based predictions~\cite{jiang2021dl}.
The grid-based prediction divides areas into equal-size grids and predicts the values for each grid (e.g., \cite{zhang2018predicting}).
However, equal-size grids are not suitable for our problem because we aim to predicate the number of available parking lots per street, which cannot divide equal-size grids.

Our work belongs to the graph-based prediction which estimates attributes of nodes on graphs (e.g., \cite{DCRNN,STGCN,wu2020connecting, wu2019graph}). 
Nodes on graphs often represent sensors such as temperature and traffic volume sensors, and their measurements are attributes of the nodes.
In our study, nodes represent parking clusters and their attributes are the parking lots availability, so the existing methods can be applied to parking predictions.
Existing methods output either a single step prediction or fixed multiple step prediction that step sizes are given in advance, and thus they do not satisfy our requirements (i.e., prediction at multiple time step without pre-defined time steps)\footnote{We note that we operated our service for several years, so it is hard to replace our model to recent models even if new models are developed.}.


\smallskip
\noindent
{\bf Graph-to-sequence models.}
Graph-to-sequence models have widely studied recently~\cite{beck2018graph,bai2019stg2seq,shen2020hierarchical}.
They generate a target of sequence from a given graph by effectively extracting information on the graph.
To the best of our knowledge, there are no works that applied to parking lots availability yet.

\section{Conclusion}
 \label{sec:conclusion}
 
We studied the parking availability prediction with SmartSantander.
We developed a graph-to-sequence neural network model to predict the number of available parking lots for the entire parking area in the whole city center and per each of the streets at multiple time steps.
We evaluated our model using a dataset containing three years of real outdoor parking data with SmartSantander, and our model achieved a highly-accurate prediction performance, which is accurate enough to develop the service in the city.
 
As part of our future works, we first plan to develop online-learning methods for capturing the current trends at real time. 
In particular, it is worthwhile to tackle drastic changes of parking, for example, due to spreading COVID-19.
Second, we plan to use real-time data through user interfaces such as user-feedback for improving the prediction performance.
Third, we plan to use alternative features, such as traffic volume and weather conditions that are collected in SmartSantander, to improve the accuracy of the parking prediction. 




%
%
%
\bibliographystyle{splncs04}
\bibliography{bibliography}
\newpage

\appendix 

\section{Baseline Sequence-to-Sequence models}

\subsubsection{RNN encoder model}
\begin{figure}[ttt]
 	\centering
 	\includegraphics[width=0.8\linewidth]{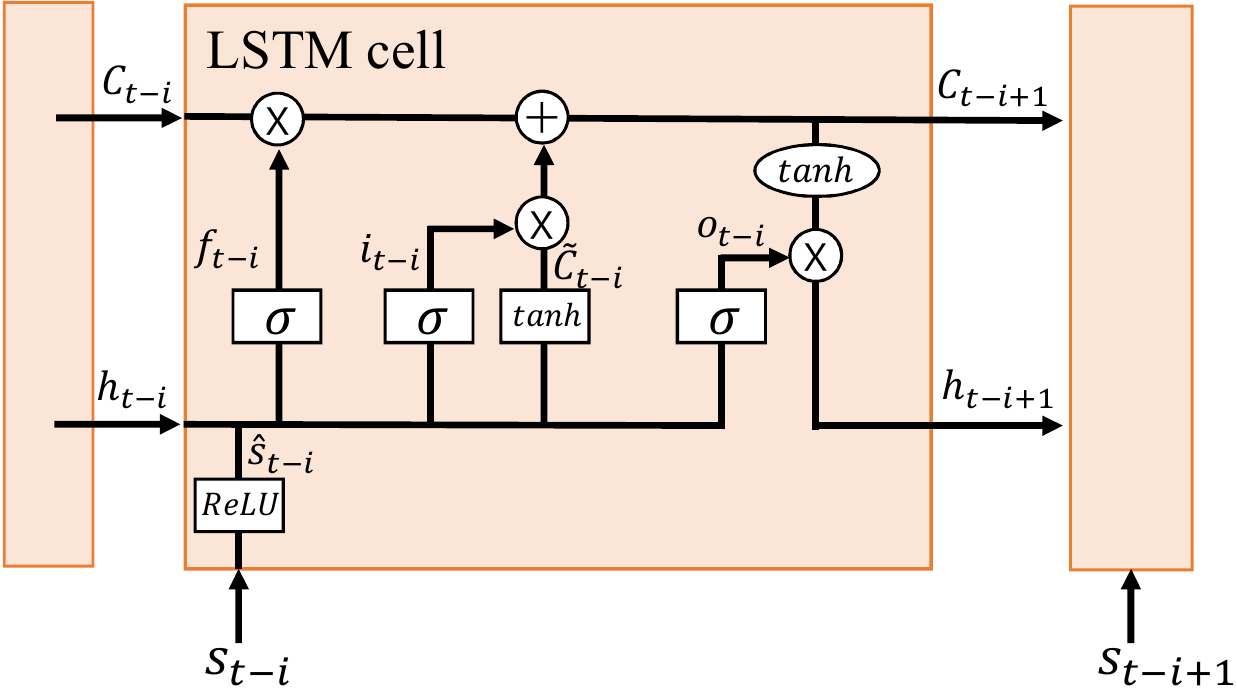}
	\caption{RNN encoder architecture}
 	\label{fig:rnn}
 \end{figure}

RNN is a neural network model whose input is a sequence of vectors, such as natural language or temporal data.
Since parking lot data is also represented as a sequence of vectors, RNN can be directly used for our problem.
The RNN encoder uses a multilayered Long Short-Term Memory (LSTM)~\cite{hochreiter1997long} whose inputs are vectors from $s_{t-M}$ to $s_{t-1}$.
 Figure \ref{fig:rnn} shows the RNN encoder.

The RNN encoder follows basic LSTM models except that we embed $s_{t-i}$ with ReLU function because it must be positive to represent the number of available parking lots.
In more concretely, the RNN encoder consists of the follow equations:
\begin{eqnarray}
\hat{\boldmath {s}}_{t} &=& ReLU (W_s  \boldmath {s}_t + \boldmath {b}_s) \label{eq:LSTMstart}\\
\boldmath {f}_{t} &=& \sigma (W_f[\boldmath{h}_{t};\hat{\boldmath{s}}_t]+\boldmath {b}_f)\nonumber\\
\boldmath {i}_{t} &=& \sigma (W_i[\boldmath {h}_{t};\hat{s}_t]+\boldmath {b}_i)\nonumber\\
\tilde{\boldmath {C}}_{t} &=& \tanh (W_C[\boldmath{h}_{t};\hat{\boldmath{s}}_t]+\boldmath {b}_C)\nonumber\\
\boldmath {C}_{t+1} &=& \boldmath {f}_t \otimes \boldmath {C}_{t-1} + \boldmath {i}_t  \tilde{C}_t\nonumber\\
\boldmath {o}_{t} &=& \sigma (W_o[\boldmath {h}_{t-1};\hat{\boldmath {s}}_t]+\boldmath {b}_o)\nonumber\\
\boldmath {h}_{t+1} &=& \boldmath {o}_t \otimes \tanh \boldmath {C}_{t+1}\nonumber
\end{eqnarray}
\noindent 
where, $W$ is weight matrix, $b$ is bias vector, and $*$ indicates Hadamard product.
$f$, $i$, $o$, $C$, and $h$ are forget gate, input gate, output gate, memory cell, and final states of hidden layer, respectively.
These equations are the same for basic LSTM cells except for Eq. (\ref{eq:LSTMstart}).

The RNN encoder can capture the temporal perspective, while it does not capture the spatial perspective at different time steps because it just inputs the vector at every single step.


 
   \begin{figure}[ttt]
 	\centering
 	\includegraphics[width=0.8\linewidth]{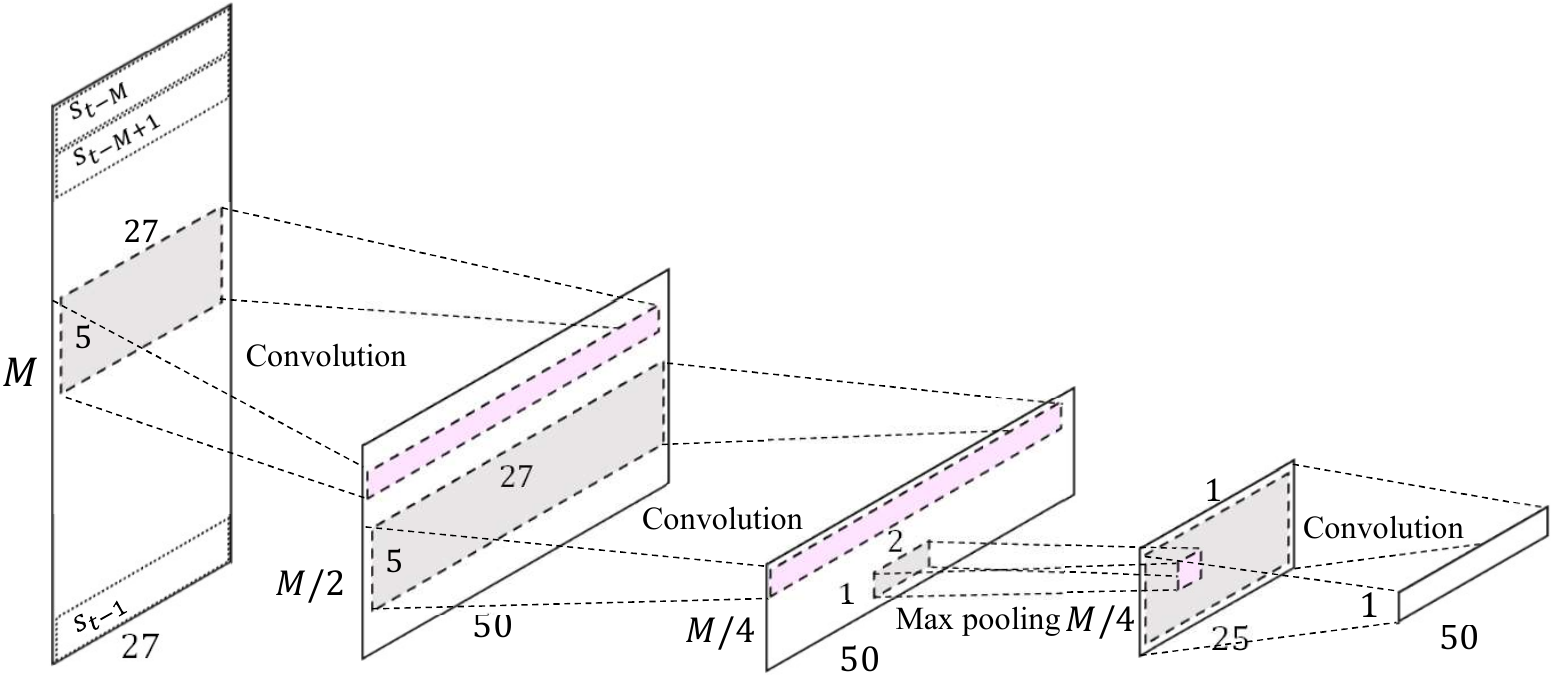}
 	\caption{CNN encoder architecture}
 	\label{fig:cnn}
 \end{figure}

\subsubsection{CNN encoder model}
CNN has been recently reported to outperform RNN regarding to the prediction of time series data \cite{tran2015learning}. 
For the CNN encoder, we employ 1D-convolution as same as our GNN encoder.
Convolutional layers output matrices that represent the characteristics of parking data with the robustness of small fluctuations.
The CNN encoder outputs the vector to the decoder, similarly to the RNN encoder model.
Figure \ref{fig:cnn} shows the CNN encoder that has three convolutional layers and one max pooling layer.
It gradually reduces the size to obtain aggregated important features.

We explain the CNN encoder in detail.
We first transform $M$ vectors (i.e., from $\boldmath{s}_{t-M}$ to $\boldmath{s}_{t-1}$) to a single matrix $\mathbb{X} \in \mathbb{R}^{M \times |s|}$, where the $i$-th column of $\mathbb{X}$ is the $s_{t-M+i-1}$. 
The design of the convolutional layer is based on the convolutional encoder architecture~\cite{zhang2017deconvolutional}.
The architecture consists of three convolutional and one max pooling layers that transform an input matrix into a fixed-length vector.
For each filter, a convolutional operation with stride length $r$ applies filter $\mathbb{W}_c \in \mathbb{R}^{f \times 50}$, where $f$ is the convolutional filter size.
Three convolutional layers have 5, 5, and $\frac{M}{4}$ as the filter sizes, and 2, 2, and $null$ as the stride length, respectively.
The reason of using $null$ is that the filter size is the same as the matrix size.
We use 50 filters in each convolutional layer.
In the CNN encoder, we also apply one dimension max pooling after the second convolutional layer to emphasize the characteristic of matrix.

The CNN encoder learns the spatial perspective from the whole city globally, but it often misses local spatial perspective that is observed a set of close parking lots.


\end{document}